\documentclass[runningheads]{llncs}
\usepackage{graphicx}
\usepackage{amsmath,amsfonts}
\usepackage{algorithmic}
\usepackage{authblk}

\usepackage{cite}
\usepackage[section]{placeins}
\usepackage{booktabs}
\usepackage{array}
\usepackage{tabularx}
\newcolumntype{L}[1]{>{\raggedright\arraybackslash}m{#1}}
\newcolumntype{C}[1]{>{\centering\arraybackslash}p{#1}}

\usepackage{marvosym}
\usepackage{amsmath} 
\usepackage{textcomp}
\pdfoutput=1

\begin{document}

 \title{User-Aware Prefix-Tuning is a Good Learner for Personalized Image Captioning}

\titlerunning{User-Aware Prefix-Tuning for Personalized Image Captioning}

\author{Xuan Wang$^{1*}$ \and
Guanhong Wang$^{1}$\thanks{Equal contribution.} \and
Wenhao Chai$^{2}$  \and \\
\vspace{-10pt}
Jiayu Zhou$^{1}$ \and
Gaoang Wang\textsuperscript{1(\Letter)}}
\authorrunning{X. Wang et al.}

\institute{Zhejiang University, Hangzhou, China \\
\email{\{xuanw,guanhongwang\}@zju.edu.cn, \{jiayu.21,gaoangwang\}@intl.zju.edu.cn}\\
\and
 University of Washington, Seattle, USA\\
\email{wchai@uw.edu}}

\maketitle              

\begin{abstract}
Image captioning bridges the gap between vision and language by automatically generating natural language descriptions for images. Traditional image captioning methods often overlook the preferences and characteristics of users. Personalized image captioning solves this problem by incorporating user prior knowledge into the model, such as writing styles and preferred vocabularies. Most existing methods emphasize the user context fusion process by memory networks or transformers. However, these methods ignore the distinct domains of each dataset. Therefore, they need to update the entire caption model parameters when meeting new samples, which is time-consuming and calculation-intensive. To address this challenge, we propose a novel personalized image captioning framework that leverages user context to consider personality factors. Additionally, our framework utilizes the prefix-tuning paradigm to extract knowledge from a frozen large language model, reducing the gap between different language domains. Specifically, we employ CLIP to extract the visual features of an image and align the semantic space using a query-guided mapping network. By incorporating the transformer layer, we merge the visual features with the user's contextual prior knowledge to generate informative prefixes. Moreover, we employ GPT-2 as the frozen large language model. With a small number of parameters to be trained, our model performs efficiently and effectively. Our model outperforms existing baseline models on Instagram and YFCC100M datasets across five evaluation metrics, demonstrating its superiority, including twofold improvements in metrics such as BLEU-4 and CIDEr.

\end{abstract}

\section{Introduction}
\label{sec:intro}

Image captioning aims to generate descriptive sentences for image content, which has drawn a lot of attention in recent years \cite{karpathy2015deep,li2023blip,wang2022image,chai2022deep}. It has a wide range of applications, such as e-commerce product description, assisting individuals with visual impairments, and automating captioning for social media posts \cite{hossain2019comprehensive,stefanini2022show}.

\begin{figure}[!t]
  \centering
  \includegraphics[width=0.95\textwidth]{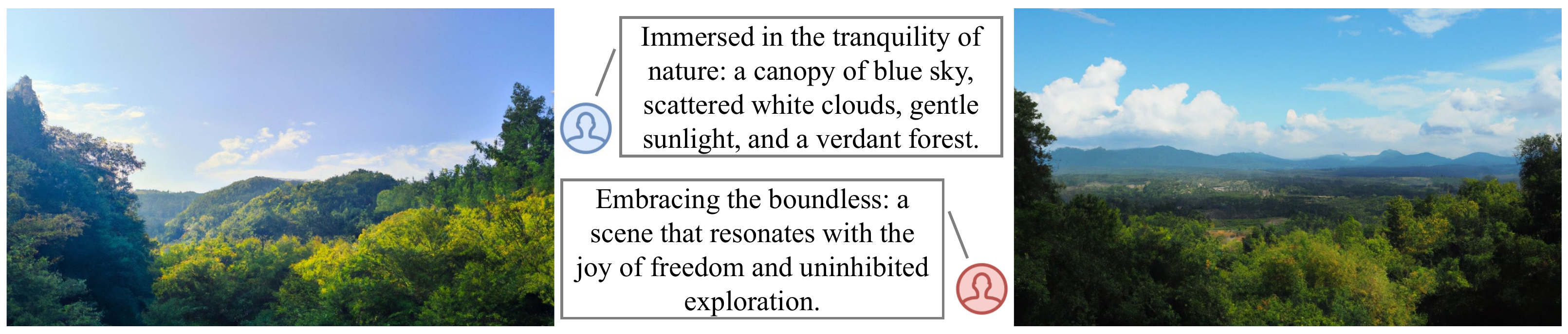}
  \caption{Owing to the personalities of the users, similar images may exhibit varying descriptions.}
  \label{fig:fig1}
\end{figure}

Previous works have explored the generation of captions using an encoder-decoder network with various attention mechanisms \cite{xu2015show,you2016image}.
However, the captions generated by these methods seem standardized and conservative in practical applications \cite{long2020cross}.
With the advent of social media, individuals have increasingly opted to express their personalities through  posting.
Fig.~\ref{fig:fig1} illustrates similar images while highlighting the contrasting captions provided by the respective users. Consequently, there is a growing need for research on personalized image captioning.
Some works \cite{park2018towards,zhang2020learning} have made early attempts to solve this problem by leveraging user context, such as active vocabularies and writing styles. These methods first encode the visual content with user information and then fuse them using memory networks or transformers to generate personalized captions. Although significant improvements have been achieved in personalized image captioning, one major issue remains in most existing methods. Considering the distinct domains of each social media dataset, these approaches require training the entire caption model for each dataset. Additionally, when encountering new samples, they need to update their caption model to generate sentences for the new data. This process requires both time and computing resources.

To address the above issue, we propose a novel personalized image captioning framework based on User-Aware Prefix-Tuning (UAPT). 
Our approach not only incorporates user-aware prior knowledge to accommodate the distinct characteristics and preferences of individuals, but also leverages the prefix-tuning paradigm \cite{li2021prefix} to exploit the linguistic capabilities of frozen large language models in order to generate precise captions, thereby bridging the gap between different domains. At first, we use the CLIP \cite{radford2021learning} image encoder to capture the visual semantics. In addition, a query-guided Mapping Network is used to align the semantic space of vision and language. 
To enable personalized characteristics of users, we employ transformer layers to fuse visual knowledge with user context prior knowledge and generate a fixed-size embedding sequence denoted as prefixes \cite{li2021prefix}. These prefixes, concatenated with caption embeddings, are subsequently fed into a GPT-2 model \cite{radford2018improving} to generate captions. Throughout this process, both the CLIP and GPT-2 models remain frozen. The Mapping Network and Fusion Network are end-to-end training, thus saving time and computing resources. The framework is shown in Fig.~\ref{fig:model}. This novel approach allows us to effectively capture the user's language style without requiring extensive personalized training data, making it more applicable in real-world scenarios.

The main contributions of this article can be summarized as follows:
\begin{itemize}
    \item To the best of our knowledge, we are the first to introduce large language models to address personalized image captioning problem with the prefix-tuning paradigm, solely training small network architecture.
    \item We propose the integration of user-aware prior knowledge with visual information to effectively capture the contextual language style of the user, enhancing personalized image captioning.
    \item Our model outperforms existing baseline models on both the Instagram and YFCC100M datasets across five evaluation metrics. Remarkably, metrics such as BLEU-4 and CIDEr show twofold improvements, providing substantial evidence of the superiority of our model.
\end{itemize}

\section{Related Work}
\label{sec:relwork}

\subsection{Image Captioning}

Image captioning serves as a pivotal link between Computer Vision (CV) and Natural Language Processing (NLP), necessitating the generation of contextually and grammatically appropriate textual descriptions for images. Notable progress has been achieved in this field with the advent of deep learning techniques. 
Earlier approaches utilized convolutional neural network \cite{krizhevsky2017imagenet} to extract visual features from images and recurrent neural network \cite{mikolov2010recurrent} to generate captions \cite{mao2014explain,vinyals2015show}.
Subsequent studies introduced attention mechanisms, leading to improved caption quality \cite{anderson2018bottom,xu2015show,you2016image}.
In addition, there has been an emergence of large-scale vision and language pre-trained models that can be applied to image captioning \cite{wang2022image,zhang2021vinvl}. However, these models often require significant computational resources and training efforts.
To address this issue, Mokady et al. \cite{mokady2021clipcap} proposed a strategy of keeping the pre-trained weights fixed and focusing on training specific network components to bridge the gap between visual and textual representations.
Despite the achievements of these efforts, they are limited to generating neutral captions, which may not fully cater to the diverse needs of users in the era characterized by the rapid emergence of social media.

\subsection{Personalized Vision and Language Research}

Driven by the rapid proliferation of social media and the growing demand for personalized content, numerous scholars have conducted research with a primary focus on customizing the visual and linguistic domains.
Shuster et al. \cite{shuster2019engaging} defined 215 potential personality traits to guide caption generation. However, in real-world scenarios, users' styles may not be adequately captured by these traits.
Park et al. \cite{park2018towards} constructed datasets specifically for personalized image captioning and employed the memory network to store the users' active vocabularies, enabling the generation of personalized captions.
Long et al. \cite{long2020cross} proposed a model for personalized image captioning that can be applied across different domains, thereby expanding the range of potential application scenarios.
Considering users' long-term and short-term stylistic preferences, Zhang et al. \cite{zhang2020learning} proposed an approach that combines these temporal dimensions. However, the implementation of this approach necessitates the collection of more detailed data.
To generate captions that align with user preferences, Wang et al. \cite{wang2018social} utilized user-provided tags as a foundational element in their methodology.
Zeng et al. \cite{zeng2019automatic} leveraged user profiles such as age and gender to generate social media comments.
Our approach eliminates the need for supplementary information, such as user profiles or posting timestamps, and only requires training a small network.

\subsection{Prompt Tuning in Image Captioning}

Prompt tuning enables large-scale pre-trained models to adapt to various tasks without requiring parameter adjustments.
Jin et al. \cite{jin2021good} conducted research demonstrating that even relatively simple hard prompts can enhance the performance of image captioning.
Ramos et al. \cite{ramos2023smallcap} utilized image-based retrieval techniques to extract relevant textual information from a database, thereby generating prompts that subsequently guided the process of caption generation.
Taking a distinct approach, Li et al. \cite{li2023blip} employed soft prompts by utilizing extracted visual representations as a condition for the large language model.
Building upon the insights from these prior works, our research extends the application of prompt tuning to the novel and challenging domain of personalized image captioning.

\section{Methods}
\label{sec:methods}

\subsection{Problem Definition}
The general personalized image captioning framework can be described as follows: Given a dataset of images, captions, and user context tuples $(x^i, c^i, u^i)$, our objective is to learn a model to generate a caption for a target input image with a specified user language style.

During training, we optimize our model in an autoregressive manner that predicts the next token without considering future tokens by minimizing negative log-likelihood loss $\mathcal{L}$, \textit{i.e.},

\begin{equation}
    \mathcal{L} = -\max _{\theta} \sum_{i=1}^{N} \sum_{j=1}^{L} \log p_{\theta}\left(c_{j}^{i} \mid x^{i},u^{i}, c_{1}^{i}, \ldots, c_{j-1}^{i}\right),
\label{eq:loss}
\end{equation}
where $\mathcal{\theta}$ denotes the model trainable parameters; $N$ is the number of samples; $L$ is the maximum length of the generated sentence.

During the process of inference, the generation of the caption is initiated by conditioning it on both the visual prefix and the user context. The subsequent tokens are then predicted sequentially, with the assistance of the language model output. At each step, the language model assigns probabilities to all tokens in the vocabulary, and these probabilities are used to determine the next token by employing a beam search algorithm.

\begin{figure}[!t]
  \centering
  \includegraphics[width=\textwidth]{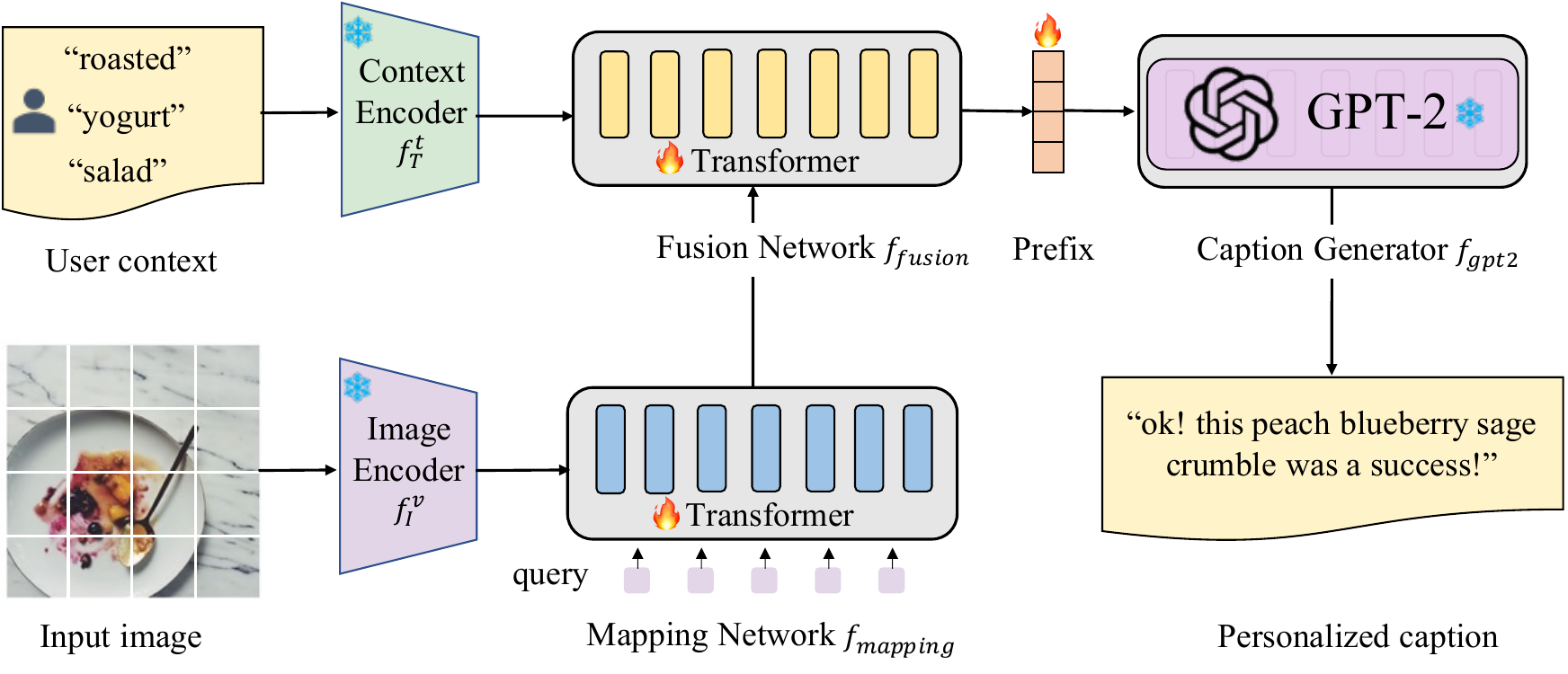}
  \caption{Overview of our User-Aware Prefix-Tuning Network (UAPT) framework. At first, we utilize a frozen image encoder $f_{I}^{v}$ and context encoder $f_{T}^{t}$ to extract visual features and user-specific embeddings, respectively. Then, a query-guided mapping network $f_{mapping}$ is exploited to align vision and language semantics. Subsequently, visual knowledge and user prior knowledge are fused by a transformer based fusion network $f_{fusion}$ to output embeddings as prefixes. Finally, the prefixes are input into a frozen large language model $f_{gpt2}$ to generate personalized captions.}
  \label{fig:model}
\end{figure}

In the subsequent section, we introduce our proposed approach, illustrated in Fig.~\ref{fig:model}. We begin by explaining the utilization of the query-guided visual knowledge mapping module to align visual cues and language semantics. Subsequently, we demonstrate the fusion of user-aware prior knowledge with visual knowledge in order to capture the contextual style of user language for personalized image captioning. Lastly, we employ prefix tuning to prompt a frozen language model for the generation of captions.

\subsection{Query-Guided Visual Knowledge Mapping}

To acquire visual cues from the images, we utilize the frozen CLIP's image encoder $f_{I}^{v}(\cdot )$ to extract visual semantics. 

Furthermore, we introduce a set of learnable query vectors $q_{i}$ to guide the Mapping Network in extracting valuable knowledge from the existing visual feature representations. This approach provides the benefit of directing the model towards extracting feature representations that are more closely aligned with the semantics of the textual space. Simultaneously, it reduces the required training resources and accelerates the training process. Consequently, the resulting mapping representation $\mathcal{V}$ can be reformulated as follows:

\begin{equation}
\begin{aligned}
    \mathcal{V} = f_{mapping} (f_{I}^{v}(x_{i} ),q_{i}).
\label{eq:mapping}
\end{aligned}
\end{equation}

\subsection{User-Aware Prior Knowledge Fusion}

Different users often exhibit unique idioms, writing styles, and other linguistic characteristics, which can be considered as personalized traits. To achieve personalized image captioning, it is crucial to incorporate user language style contexts into the network. In this regard, we utilize TF-IDF denoted as $f_{tf-idf}(\cdot)$ to represent the users' distinctive writing habits, which can measure the importance of a word in a document \cite{ramos2003using}.
Specifically, we consider each user's historical postings as a document and apply the TF-IDF to extract personalized keywords for each user. These keywords capture the unique vocabulary and writing style employed by the users in their posts. Subsequently, we input these obtained keywords into the Context Encoder, instantiated as CLIP's text encoder $f_{T}^{t}(\cdot)$, to encode them and generate a distinctive embedding vector for each user.

Furthermore, we employ a transformer $f_{fusion}(\cdot)$ to fuse the user's embedded representation with the visual projection representation, enabling a comprehensive interaction between the image information and the user's linguistic style context. This fusion facilitates a stronger correlation between the semantic information of the image and the user's style.
The final fusion vector, represented as the output $\mathcal{P}$, can be mathematically expressed as follows:

\begin{equation}
\begin{aligned}
    \mathcal{P}  = f_{fusion}( f_{T}^{t}(f_{tf-idf}(u_{i} ) ),\mathcal{V} ).
\label{eq:fusion}
\end{aligned}
\end{equation}

\subsection{Caption Generation}

During the training process, we employ a pre-trained GPT-2 model to generate image captions. Specifically, we input the fusion vectors as prefixes to the GPT-2 model, allowing the model to learn the semantic information of the images and the writing habits of the users to generate the corresponding captions.
In the inference process, we utilize the beam search algorithm to generate more accurate and coherent captions. 
The entire process can be summarized as follows:

\begin{equation}
\begin{aligned}
    \mathcal{C} = f_{gpt2}  (\mathcal{P} ).
\label{eq:decoder}
\end{aligned}
\end{equation}

\section{Experiments}
\label{sec:exp}

\subsection{Dataset}
\label{subsec:dataset}

\begin{table*}[!t]
\begin{center}
\caption{Statistics of the datasets.} 
\label{tab:table1}

\resizebox{0.99\textwidth}{!}{ 
\begin{tabular}{L{2.2cm}|C{1.4cm}C{1.4cm}C{1.8cm}C{2.0cm}|C{1.8cm}C{1.8cm}}
  \toprule
  \textbf{Datasets} &\textbf{Posts} &\textbf{Users} &\textbf{Posts/User} &\textbf{Words/Post} &\textbf{Intra-class} &\textbf{Inter-class} \\
  \midrule
  Instagram &721,176 &4,820 &149.6 &8.55 &0.0330 &0.0088  \\
  YFCC100M  &462,036 &6,197 &74.6 &6.30 &0.1682 &0.0044  \\
  \bottomrule
\end{tabular}
}
\end{center}
\end{table*}

We validate the performance of our model using two datasets \cite{park2018towards}: Instagram and YFCC100M. The datasets are partitioned following the methodology outlined in \cite{park2018towards}. Detailed statistics regarding the datasets are presented in Table~\ref{tab:table1}.

To assess the feasibility and indispensability of personalization, we treat each user as a separate class and evaluate the intra-class and inter-class similarity of the captions. The results of this analysis, presented in the right part of Table \ref{tab:table1}, unequivocally demonstrate the existence of distinct textual preferences among individual users. This compelling evidence further supports the idea that user characteristics can be effectively captured by considering their posts.

\begin{table*}[!t]
\centering
\caption{Results of evaluation metrics on the Instagram and YFCC100M datasets. The best results for each evaluation metric are marked in bold.}
\label{tab:metric}
\resizebox{0.99\textwidth}{!}{
\begin{tabular}{L{2.2cm}|C{1.4cm}C{1.4cm}C{1.4cm}C{1.4cm}|C{1.6cm}C{1.8cm}C{1.2cm}}
  \toprule
  \multicolumn{8}{c}{\textbf{Instagram}} \\
  \midrule
  \textbf{Methods} &\textbf{BLEU-1} &\textbf{BLEU-2} &\textbf{BLEU-3} &\textbf{BLEU-4} &\textbf{METEOR} &\textbf{ROUGE-L} &\textbf{CIDEr} \\
  \midrule
  ShowTell  &0.055 &0.019 &0.007 &0.003 &0.038 &0.081 &0.004 \\
  ShowAttTell &0.106 &0.015 &0.000 &0.000 &0.026 &0.026 &0.049 \\
  1NN-Im &0.071 &0.020 &0.007 &0.004 &0.032 &0.069 &0.059  \\
  \midrule
  Seq2seq &0.050 &0.012 &0.003 &0.000 &0.024 &0.065 &0.034 \\
  1NN-Usr &0.063 &0.014 &0.002 &0.000 &0.028 &0.059 &0.025  \\
  1NN-UsrIm &0.106 &0.032 &0.011 &0.005 &0.045 &0.104 &0.084  \\
  CSMN-P5 &\textbf{0.171} &0.068 &0.029 &0.013 &0.064 &\textbf{0.177} &0.214  \\
  CSMN-P5-Mul &0.145 &0.049 &0.022 &0.009 &0.049 &0.145 &0.143  \\
  \midrule
  Ours &0.161  &\textbf{0.084} &\textbf{0.050} &\textbf{0.032} &\textbf{0.097} &0.160 &\textbf{0.343}  \\
  \midrule
  \midrule
  \multicolumn{8}{c}{\textbf{YFCC100M}} \\
  \midrule
  \textbf{Methods} &\textbf{BLEU-1} &\textbf{BLEU-2} &\textbf{BLEU-3} &\textbf{BLEU-4} &\textbf{METEOR} &\textbf{ROUGE-L} &\textbf{CIDEr} \\
  \midrule
  ShowTell  &0.027 &0.003 &0.000 &0.000 &0.024 &0.043 &0.003 \\
  ShowAttTell &0.088 &0.010 &0.001 &0.000 &0.034 &0.116 &0.076 \\
  1NN-Im &0.033 &0.006 &0.001 &0.000 &0.020 &0.043 &0.063  \\
  \midrule
  Seq2seq &0.076 &0.010 &0.000 &0.000 &0.034 &0.066 &0.069 \\
  1NN-Usr &0.032 &0.003 &0.001 &0.000 &0.016 &0.051 &0.028  \\
  1NN-UsrIm &0.039 &0.005 &0.001 &0.000 &0.021 &0.050 &0.050  \\
  CSMN-P5 &0.106 &0.034 &0.012 &0.004 &0.033 &0.099 &0.064  \\
  CSMN-P5-Mul &\textbf{0.116} &0.036 &0.010 &0.003 &0.036 &\textbf{0.111} &0.060  \\
  \midrule
  Ours &0.100  &\textbf{0.042} &\textbf{0.018} &\textbf{0.008} &\textbf{0.064} &0.097 &\textbf{0.170}  \\
  \bottomrule
\end{tabular}
}
\end{table*}

\subsection{Implementation Details}
\label{subsec:implementation}
In the UAPT, both the Mapping Network and the Fusion Network consist of $5$ layers of multi-head self-attention with $8$ heads each.
The lengths of the learnable query, visual feature representations, and user's embedding vector are $16, 16, 4$ respectively.
For the Image Encoder and Context Encoder, we utilize the pre-trained image encoder and text encoder from EVA-CLIP \cite{sun2023eva}. Additionally, GPT-2 (medium) \cite{radford2019language} is employed as the Caption Generator.

We conduct training for $6$ epochs using a batch size of $50$ on a single NVIDIA GeForce RTX3090. The AdamW \cite{loshchilov2017decoupled} optimizer is employed with $\beta_1=0.9$, $\beta_2=0.96$, and a weight decay rate of $0.005$. To schedule the learning rate, we utilize cosine learning rate decay with a peak learning rate of $6e-4$, accompanied by linear warm-up over $6k$ steps. 
During the inference phase, the temperature of the beam search is set to $0.8$, and the beam size is set to $3$.

\begin{table*}[!t]
\begin{center}
\caption{Ablation study on the YFCC100M. The best results for each evaluation metric are marked in bold.} 
\label{tab:table4}

\resizebox{0.99\textwidth}{!}{ 
\begin{tabular}{L{2.2cm}|C{1.4cm}C{1.4cm}C{1.4cm}C{1.4cm}|C{1.6cm}C{1.8cm}C{1.2cm}}
  \toprule
  \textbf{Methods} &\textbf{BLEU-1} &\textbf{BLEU-2} &\textbf{BLEU-3} &\textbf{BLEU-4} &\textbf{METEOR} &\textbf{ROUGE-L} &\textbf{CIDEr} \\
  \midrule
  w/o Context &0.074 &0.030 &0.013 &0.006 &0.052 &0.075 &0.121 \\
  w/o Mapping &0.083 &0.034 &0.015 &0.007 &0.060 &0.088 &0.154 \\
  w/o Fusion &0.085 &0.036 &0.016 &0.007 &0.061 &0.088 &0.161  \\
  w/o Query &0.083 &0.035 &0.015 &0.006 &0.061 &0.089 &0.152 \\
  \midrule
  Full Model &\textbf{0.100}  &\textbf{0.042} &\textbf{0.018} &\textbf{0.008} &\textbf{0.064} &\textbf{0.097} &\textbf{0.170}  \\
  \bottomrule
\end{tabular}
}
\end{center}
\end{table*}

\subsection{Baselines}
\label{subsec:baselines}

To demonstrate the effectiveness of our model, we select several baseline models for comprehensive comparative analysis, including:
1) \textbf{ShowTell} \cite{vinyals2015show}: ShowTell is an encoder-decoder model designed for caption generation, employing an RNN decoder.
2) \textbf{ShowAttTell} \cite{xu2015show}: ShowAttTell integrates visual attention computation to capture the importance of individual image regions during word decoding.
3) \textbf{1NN-Im}, \textbf{1NN-Usr}, and \textbf{1NN-UsrIm} \cite{park2018towards}: These baselines employ a retrieval-based approach by utilizing the captions from the nearest training image, nearest user, or a combination of both as the basis for generating captions.
4) \textbf{Seq2seq} \cite{vinyals2015grammar}: Seq2seq is a recursive neural network with three hidden LSTM layers. This baseline retrieves 60 active words from the querying user in a descending order of TF-IDF weights and predicts the corresponding caption.
5) \textbf{CSMN-P5} and \textbf{CSMN-P5-Mul} \cite{park2018towards}: These baselines integrated the user's active vocabulary as part of their memory network. CSMN-P5-Mul extended CSMN-P5 by incorporating multi-layer CNNs into its architecture.

\subsection{Quantitative Analysis}
\label{subsec:quan}

We utilize BLEU \cite{papineni2002bleu}, METEOR \cite{banerjee2005meteor}, ROUGE-L \cite{lin2004rouge}, and CIDEr \cite{vedantam2015cider} as evaluation metrics. 
Table~\ref{tab:metric} presents the performance of UAPT and the compared baselines on the two datasets. The first part of the table corresponds to architectures without personalization, while the remaining results incorporate personalization.
Across both datasets, our model outperforms the existing baseline models across five evaluation metrics. In the Instagram dataset, our model achieves significant improvements, surpassing other baseline models by more than double in metrics such as BLEU-3, BLEU-4, and CIDEr. Similarly, in the YFCC100M dataset, our model demonstrates substantial enhancements, more than doubling the performance of the baseline models in terms of BLEU-4 and CIDEr scores. These findings conclusively demonstrate the superior performance of our model.

\subsection{Ablation Study}
\label{subsec:ablation}

We perform ablation experiments to evaluate the effectiveness of each component in the model, as shown in Table \ref{tab:table4}. These experiments included the following variations: 1) \textbf{``w/o Context"} which removes the Context Encoder; 2) \textbf{``w/o Mapping"} which eliminates the Mapping Network; 3) \textbf{``w/o Fusion"} which excludes the Fusion Network; 4) \textbf{``w/o Query"} where the value of learnable query is set to $0$ and the weights are frozen.

The exclusion of any component from the model results in a decline in the evaluation metrics, highlighting the indispensability of each constituent. Among the variations, the impact of ``w/o Context" is the most pronounced, emphasizing the importance of user-aware prior knowledge in generating higher-quality captions. This finding further supports the notion that our model successfully constructs unique embedding sequences for each user, enabling accurate guidance in generating personalized captions.
\begin{figure}[!t]
  \centering
  \includegraphics[width=\textwidth]{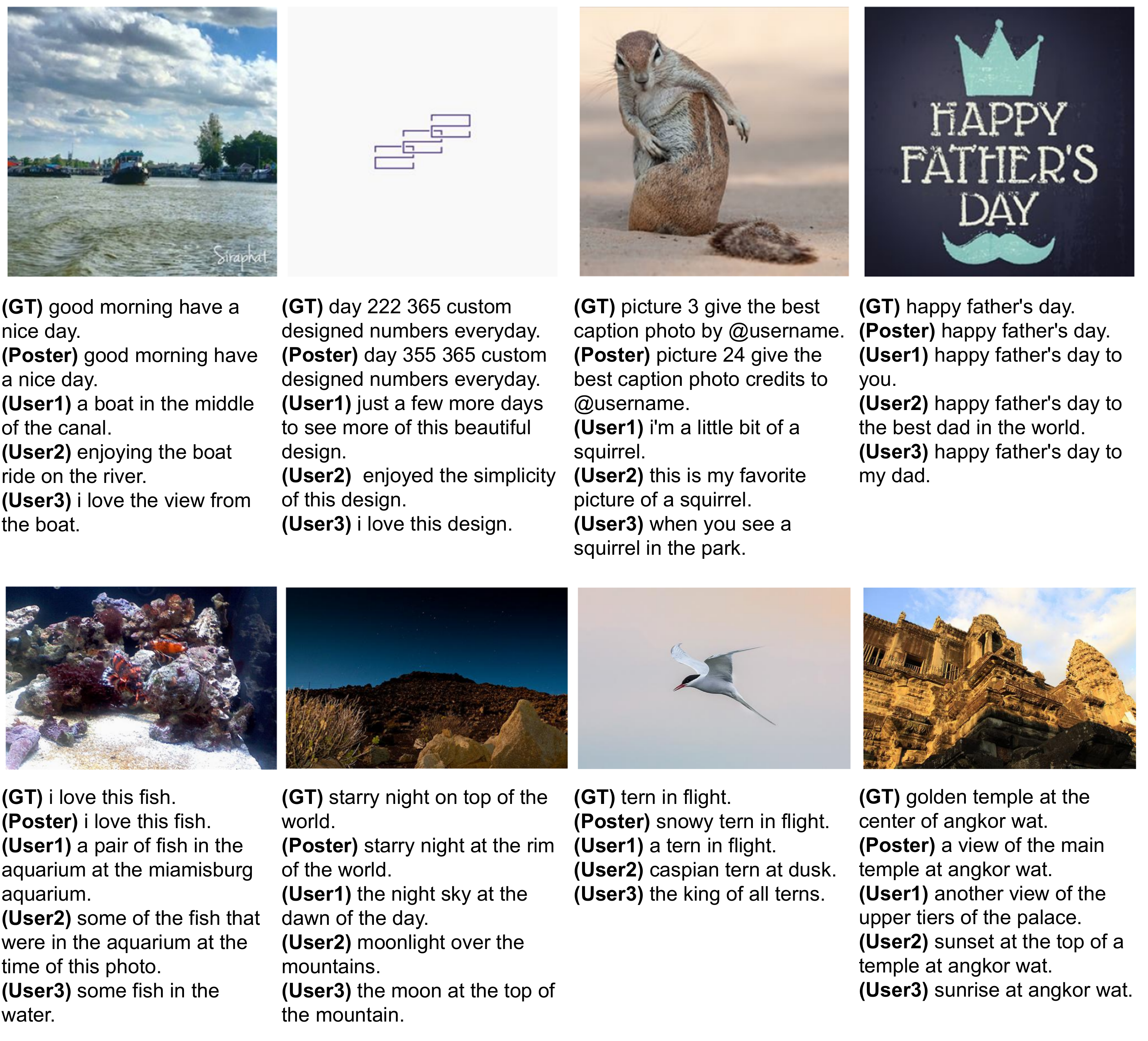}
  \caption{Examples from Instagram (top) and YFCC100M (bottom).}
  \label{fig:ins}
\end{figure}

\subsection{Qualitative Analysis}
\label{subsec:qual}

Fig. \ref{fig:ins} presents authentic examples of generated captions from the Instagram and YFCC100M datasets. Each instance includes the ground truth (GT) caption, captions generated by the original posting user (Poster) and other users (User1, User2, and User3).

By employing our model, we can generate captions for Posters that are either identical or remarkably similar to the GT. It is noteworthy that even in situations where the correlation between images and text is relatively weak, as demonstrated in the second and third examples, our model can still generate captions that align with the users' preferences by acquiring knowledge of their writing patterns.
Moreover, these examples reveal user-specific vocabulary tendencies. For example, User2 often uses ``enjoy" in captions, and User3 frequently expresses emotions through ``i love".
This demonstrates our model's proficiency in capturing users' characteristics and generating captions that align with their preferences and styles. Moreover, different users produce captions in their distinctive styles for the same image, further substantiating our model's personalized generation capability.

In the YFCC100M dataset, it is noteworthy that our model demonstrates the capability to generate captions that exhibit a higher level of precision and specificity compared to the GT, as exemplified in the third example. This observation underscores the remarkable performance and potential of our model.
\section{Conclusions}
\label{sec:conc}
In this paper, we have proposed a novel personalized image captioning framework based on user-aware prefix-tuning. Our method utilizes query-guided visual knowledge mapping for aligning vision and language semantics, while also fusing user-aware prior knowledge to consider the characteristics and preferences of users. Finally, we leverage prefix-tuning to extract knowledge from frozen large language model, narrowing the gap between different semantic domains. With a small number of parameters to be trained, our model boosts the performance efficiently and effectively.
The experiments conducted on two publicly available datasets demonstrate the superiority of our model and validate the contributions of our method.

In the future, we will explore how to integrate more fine-grained user context to generate more captivating and tailored descriptions for personalized image captioning.

The Gaussian Mixture Model (GMM) is a traditional machine learning model used for tasks such as probabilistic modeling and clustering. Its network structure is relatively straightforward, typically consisting of multiple Gaussian distributions representing different data clusters or categories. During training, the GMM model estimates the parameters (mean and variance) of each Gaussian distribution and the probabilities of data points belonging to each cluster. In image segmentation tasks, the GMM model can be employed to classify image pixels into various categories, such as background and low-grade gliomas. The GMM model offers advantages such as computational efficiency and interpretability, but its effectiveness in modeling complex data with high-dimensional features is limited.

To further elaborate on the GMM model, let's consider its application in the context of the LGG-MRI-Segmentation dataset. In this case, the GMM is used to model the distribution of pixel intensities in the MRI images. Each Gaussian component in the GMM represents a distinct tissue class or region, such as tumor or normal brain tissue.

The probability density function of the GMM is defined as a weighted sum of K Gaussian densities:

$p(x|\theta) = \sum_{k=1}^K \pi_k \mathcal{N}(x|\mu_k, \Sigma_k)$

where $x$ is the feature vector of a pixel, $\pi_k$ is the mixing coefficient (weight) of the k-th Gaussian component, and $\mathcal{N}(x|\mu_k, \Sigma_k)$ is the Gaussian density function with mean $\mu_k$ and covariance matrix $\Sigma_k$.

The parameters of the GMM are estimated using the Expectation-Maximization (EM) algorithm, which iteratively alternates between two steps:

E-step: Calculate the posterior probabilities (responsibilities) of each pixel belonging to each Gaussian component based on the current model parameters.
M-step: Update the model parameters (means, covariances, and mixing coefficients) to maximize the likelihood of the data given the current responsibilities.
Once the GMM is trained, it can be used to segment new MRI images by assigning each pixel to the Gaussian component with the highest posterior probability.

While the GMM is a simple and interpretable model, it has some limitations when applied to complex datasets like LGG-MRI-Segmentation. The assumption of Gaussian distributions for each tissue class may not always hold true, especially in the presence of artifacts, noise, or heterogeneous tumor appearances. Moreover, the GMM does not explicitly model the spatial dependencies between neighboring pixels, which can lead to fragmented or noisy segmentations.

To address these limitations, various extensions and improvements to the GMM have been proposed, such as incorporating spatial priors, using non-Gaussian distributions, or combining the GMM with other models like Markov Random Fields. However, in recent years, deep learning-based approaches have largely outperformed traditional methods like GMM for medical image segmentation tasks, thanks to their ability to learn complex, hierarchical features from large datasets.

\section*{Acknowledgments}
This work is supported by the Fundamental Research Funds for the Central Universities (No.226-2023-00045) and National Natural Science Foundation of China (No.62106219).

\bibliographystyle{splncs04}
\bibliography{reference}

\end{document}